\documentclass[11pt,letterpaper]{article}

\usepackage{emnlp2017}
\usepackage{times}
\usepackage{latexsym}

\usepackage{amsfonts}
\usepackage{amsmath}
\usepackage{amssymb}
\usepackage{footnote}

\usepackage[T1]{fontenc}
\usepackage{CJKutf8}
\usepackage[english]{babel}

\usepackage{makecell}
\usepackage{bookmark}
\usepackage{booktabs}
\usepackage{graphicx}
\usepackage{diagbox}
\usepackage{subcaption}
\usepackage[export]{adjustbox}

\emnlpfinalcopy

\def\emnlppaperid{***}

\title{A Syllable-based Technique for Word Embeddings of Korean Words}

\author{
    Sanghyuk Choi\thanks{$\,\,$ Portions of this research were done while the author was a student at Seoul National University.} \and Taeuk Kim \and Jinseok Seol \and Sang-goo Lee \\
    Department of Computer Science and Engineering \\
    Seoul National University \\
    {\tt \{sanghyuk, taeuk, jamie, sglee\}@europa.snu.ac.kr}
}

\date{}

\begin{document}

\maketitle

\begin{abstract}
Word embedding has become a fundamental component to many NLP tasks such as named entity recognition and machine translation.
However, popular models that learn such embeddings are unaware of the morphology of words, so it is not directly applicable to highly agglutinative languages such as Korean.
We propose a syllable-based learning model for Korean using a convolutional neural network, in which word representation is composed of trained syllable vectors.
Our model successfully produces morphologically meaningful representation of Korean words compared to the original Skip-gram embeddings. The results also show that it is quite robust to the Out-of-Vocabulary problem.
\end{abstract}

\section{Introduction}

Continuous word representation has been a fundamental ingredient to many NLP tasks with the advent of simple and successful approaches such as Word2Vec \cite{mikolov2013efficient} and GloVe \cite{pennington2014glove}.
Although it has been verified that they are effective in formulating semantic and syntactic relationship between words, there are some limitations.
First, they are only available to words in pre-defined vocabulary thus prone to the Out-of-Vocabulary(OOV) problem.
Second, they cannot utilize subword information at all because they regard word as a basic unit.
Those problems become more magnified when applying word-based methods to agglutinative languages such as Korean, Japanese, Turkish, and Finnish.
In this work, we propose a new model that utilizes syllables as basic components of word representation to alleviate the problems, especially for Korean.
In our experiment, we confirm that our model constructs  representation of words which contains a semantic and syntactic relationship between words.
We also show that our model can handle OOV problem and capture morphological information without dedicated analysis.

\section{Related Work}

Recent works that utilize subword information to construct word representation could be largely divided into two families: The models that use morphemes as a component and the others taking advantage of characters.

\vspace{2mm}
\noindent\textbf{Morpheme-based representation models}

A morpheme is the smallest unit of meaning in linguistics.
Therefore, there are many researches that consider morphemes when building word representations \cite{luong2013better, botha2014compositional, cotterell2015morphological}.

\newcite{luong2013better} applies a recursive neural network over morpheme embeddings to obtain word embeddings.
Although morpheme-based models are good at capturing semantics, one major drawback is that most of them require manually annotated data or an explicit morphological analyzer which could introduce unintended errors.
Our model doesn't need such a preprocessing.

\begin{figure*}[!htbp]
	\includegraphics[width=1.0\textwidth,height=200pt]{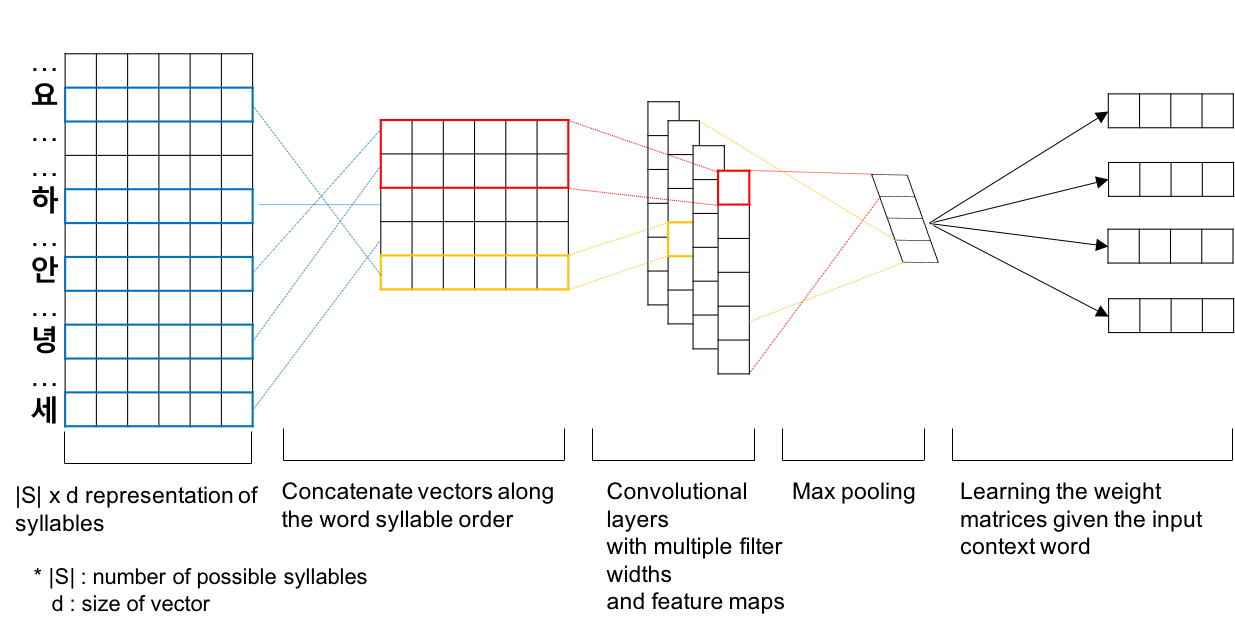}
	\caption{Overall architecture of our model. Each syllable is a $d$-dimensional vector. For a given word `\begin{CJK}{UTF8}{mj}안녕하세요\end{CJK}' (hello, \emph{annyeonghaseyo}), we concatenate vectors according to syllable order in word. After passing through the convolutional layer and max pooling layer, word representation is produced. All parameters are jointly trained by Skip-gram scheme.\label{fig:model}}
\end{figure*}

\vspace{2mm}
\noindent\textbf{Character-based representation models}

Recently, utilizing information from characters has become one of the active NLP research topics.
One way to extract knowledge from a sequence of characters is using character n-grams \cite{wieting2016charagram, bojanowski2016enriching}.

\newcite{bojanowski2016enriching} suggests an approach based on the Skip-gram model \cite{mikolov2013efficient}, where the model sums character n-gram vectors to represent a word.
On the other hand, there are some approaches \cite{dos2014deep, ling2015finding, santos2015boosting, zhang2015character, kim2016character, jozefowicz2016exploring, chung2016character} in which word representations are composed of character embeddings via deep neural networks such as convolutional neural networks (CNN) or recurrent neural networks (RNN).

\newcite{kim2016character} introduces a language model that aggregates subword information through a character-level CNN. 
Models based on characters have shown competitive results on many tasks.
A problem of character-based models is that characters themselves have no semantic meanings so that models often concentrate on only local syntactic features of words.
To avoid the problem, we select syllables which have fine-granularity like a character but has its own meaning in Korean as a basic component of the representation of words.

\section{Proposed Model}

\noindent\textbf{Characteristics of Korean Words}

Morphologically, unlike many other languages, a Korean word (\emph{Eojeol}) is not just a concatenation of characters.
It is constructed by the following hierarchy: a sequence of syllables (\emph{Eumjeol}) forms a word, and the composition of 2 or 3 characters (\emph{Jaso}) forms a syllable \cite{kang1994syllable}.

In linguistics, Korean language is categorized as an agglutinative language, where each word is made of a set of morphemes.
To complete the Korean word (\emph{Eumjeol}), a root morpheme must be combined with a bound morpheme (\emph{Josa}), or a postposition (\emph{Eomi}). 
This derivation produces about 60 different forms of the similar meaning, which causes the explosion of vocabulary.
For the same reason, the number of occurrences of each word is relatively small even with a large corpus, which prevents the model from an efficient learning.
Thus, most of the Korean word representation models use morphemes as an embedding unit, though it requires an additional preprocessing.
The problem is that errors coming from an immature morpheme analyzer might be propagated to the word representation model.
Moreover, a single Korean syllable possess a semantic meaning.
For example, the word `\begin{CJK}{UTF8}{mj}대학\end{CJK}'(college, \emph{daehag}) is a composition of `\begin{CJK}{UTF8}{mj}대\end{CJK}'(big, or great, \emph{dae}) and `\begin{CJK}{UTF8}{mj}학\end{CJK}'(learn, or a study, \emph{hag}).
Therefore, our model regards syllables as embedding units rather than words or morphemes.
For instance, the representations of `\begin{CJK}{UTF8}{mj}나는\end{CJK}'(I am, \emph{naneun}), `\begin{CJK}{UTF8}{mj}나의\end{CJK}'(my, \emph{naui}), or `\begin{CJK}{UTF8}{mj}나에게\end{CJK}'(to me, \emph{na-ege}) are constructed by leveraging the same syllable vector `\begin{CJK}{UTF8}{mj}나\end{CJK}'(I, \emph{na}).

\vspace{5mm}
\noindent\textbf{Syllable-based Representation}

Similar to \cite{kim2016character}, let $\mathcal{S}$ be a set of all Korean syllables.
We embed each syllables into $d$-dimensional vector space, so that $Q \in \mathbb{R}^{d \times |\mathcal{S}|}$ becomes a syllable embedding matrix.
Let $(s_1, s_2 , ..., s_l)$ denote a word $t \in V$ which consists of $l$ syllables, $t$ is represented by concatenating syllable vectors as a column vector: $(Qs_1, Qs_2, ..., Qs_l) \in \mathbb{R}^{d \times l}$.
Then we apply a convolution filter $H \in \mathbb{R}^{d \times w}$ having a width $w$, we get a feature map $f^t \in \mathbb{R}^{l-w+1}$.
For filters whose widths are more than 1, they need a zero padding when processing words coming from only a single syllable.

In detail, for the given filter $H$, the feature map can be calculated as follows:
\begin{equation}
f^t_i = \tanh (\langle (Qs_i, ..., Qs_{i+w-1}) ,H \rangle + b)
\end{equation}
\noindent where $\langle A, B \rangle = \text{tr}(AB^\intercal)$ denoting Frobenius inner product.
We then apply a max pooling $y^t = \max_i f^t_i$ to extract the most important feature.
By using multiple filters, namely $H_1, H_2, ..., H_h$, we get a final representation $y^t = (y^t_1, ..., y^t_h)$ for the word $t$.

For training, we adopt Skip-gram \cite{mikolov2013distributed} method with negative sampling so that for a given center word $y^t$, we maximize the log-probability of predicting context word $y^c$.
We jointly train syllable embedding matrix and convolution filters all together.
Figure \ref{fig:model} shows overall architecture of our model.

\section{Experiments and Results}

\noindent\textbf{Datasets and Baselines}

The Experiments are performed on a randomly sampled subset of Korean News corpus collected from 2012 to 2014, containing approximately 2.7M tokens, 11k vocabulary, and 1k syllables.
We compare our model to the original \emph{skip-gram model with negative sampling}~\cite{mikolov2013distributed} as a baseline.

\vspace{2mm}
\noindent\textbf{Implementation details}

For all experiments, we use the following common parameters for both our model and baseline.
We use vector representations of dimension 320, the size of window is 4 and the negative-sampling parameter is 7.
We train over twelve epochs.
In our model, the dimension of syllable embedding is 320.
Empirically, using filters with size 1\textasciitilde4 was enough since most of Korean words are composed of 2\textasciitilde4 syllables\footnote[1]{About 95\% of words in a training set had a length less than 5.}.

\begin{figure}[!htbp]
  \includegraphics[width=\linewidth]{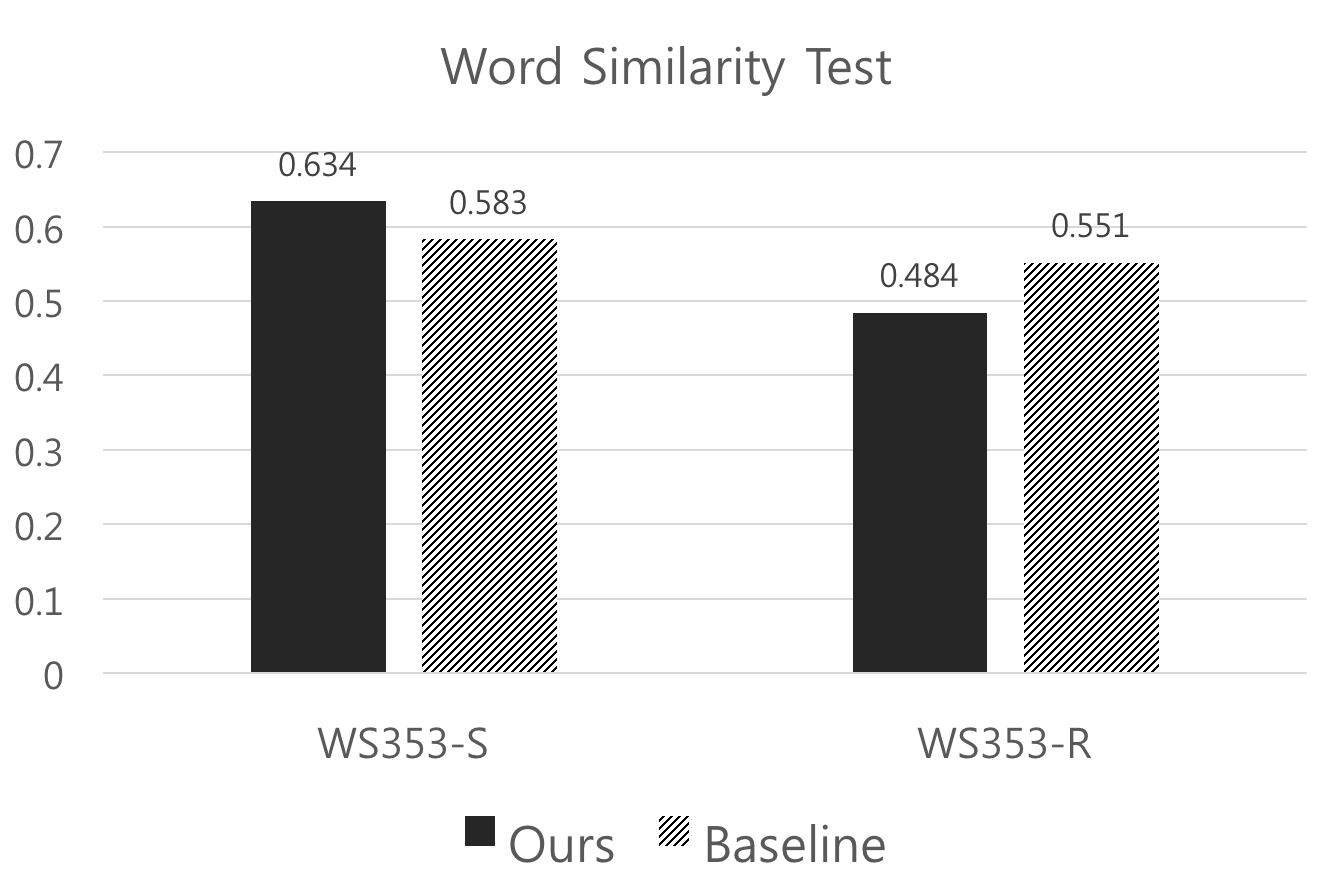}
  \caption{Test result on translated WordSim353 dataset. It contains similarity and relatedness test and measured by Pearson correlation. Our model outperformed the baseline in similarity task.\label{fig:score}}
\end{figure}

\subsection{Quantitative Evaluation}

We use the WordSim353 dataset \cite{finkelstein2001placing, agirre2009study} for the word similarity and relatedness task.
As WordSim353 dataset is an English data, we translated it into Korean.
The quality of the word vector representation is evaluated by computing Pearson correlation coefficient between human judgment scores and the cosine similarity between word vectors.

The graph in Figure \ref{fig:score} shows that our model outperforms the baseline on WS353-Similarity dataset.
We estimated it since a lot of similar words share the same syllable(s) in Korean.
On the other hand, on WS353-Relatedness, the performance is not as good in comparison with the similarity task.
We presume that leveraging syllables on computing representations can be a noise among related words without common syllables.

\subsection{Qualitative Evaluation}

\noindent\textbf{Out-Of-Vocabulary Test}

\begin{table}
  \centering
  \begin{tabular}{c|c}
    \toprule
    Original word & Newly coined word \\
    \midrule
    \makecell{\begin{CJK}{UTF8}{mj}구글\end{CJK}\\(Google, \emph{gugeul})} & \makecell{\begin{CJK}{UTF8}{mj}구글신\end{CJK}\\(God google, \emph{gugeulsin})} \\
    \makecell{\begin{CJK}{UTF8}{mj}이득\end{CJK}\\(Profit, \emph{ideug})} & \makecell{\begin{CJK}{UTF8}{mj}개이득\end{CJK}\\(Real profit, \emph{gaeideug})} \\
    \makecell{\begin{CJK}{UTF8}{mj}퇴근\end{CJK}\\(Leave work,\\ \emph{toegeun})} & \makecell{\begin{CJK}{UTF8}{mj}퇴근각\end{CJK}\\(Time to leave work,\\ \emph{toegeungag})} \\
    \makecell{\begin{CJK}{UTF8}{mj}갤럭시노트\end{CJK}\\(Galaxy Note,\\ \emph{gaelleogsinoteu})} & \makecell{\begin{CJK}{UTF8}{mj}갤노트\end{CJK}\\(Gal'Note,\\ \emph{gaelnoteu})} \\
    \bottomrule
  \end{tabular}
  \caption{4 newly coined words in Korean which did not appear in training data. Proposed model successfully recognized stem from the original word, and predicted it as the most similar word.\label{tab:oov}}
\end{table}

Since our model uses syllable vectors when computing word representation, it is possible to achieve representation of OOV words by combining syllables.
To evaluate the representations of OOV words, we manually chose 4 newly coined words not appear in training data (Table \ref{tab:oov}).
These words were derived from original words.
For example, `\begin{CJK}{UTF8}{mj}구글신\end{CJK}'(God Google, \emph{gugeulsin}) is derived from `\begin{CJK}{UTF8}{mj}구글\end{CJK}'(Google, \emph{gugeul}) and `\begin{CJK}{UTF8}{mj}갤노트\end{CJK}`(Gal'Note, \emph{gaelnoteu}) is a abbreviation form of `\begin{CJK}{UTF8}{mj}갤럭시노트\end{CJK}'(Galaxy Note, \emph{gaelleogsinoteu}).
Morphologically, two of them concatenate additional syllables to the original word, and the other two remove some syllables.

We examined the nearest neighbor of the representations of OOV words, and confirmed that each original word vector is placed in the nearest distance.
It is no wonder since almost every newly coined word keeps the syllables of original word with their positions fixed.

\vspace{2mm}
\noindent\textbf{Morphological Representation Test}

We now evaluate our model on language morphology by observing how word representation leverages morphological characteristics.
As mentioned above, the process of forming a sentence of Korean is totally different from many other languages.
In case of Korean, a word can function in the sentence only if it is combined with the bound morpheme.
For example, `\begin{CJK}{UTF8}{mj}서울을\end{CJK}'(of Seoul, \emph{seoul-eul}) is a combination of full morpheme `\begin{CJK}{UTF8}{mj}서울\end{CJK}'(Seoul, \emph{seoul}) + bound morpheme `\begin{CJK}{UTF8}{mj}을\end{CJK}'(of, \emph{eul}).

\begin{figure}[!ht]
  \adjustbox{minipage=2em,valign=c}{\subcaption{}\label{sfig:pca-a}}%
  \begin{subfigure}[c]{\dimexpr0.9\linewidth-1em\relax}
    \centering
    \includegraphics[width=\textwidth,valign=c]{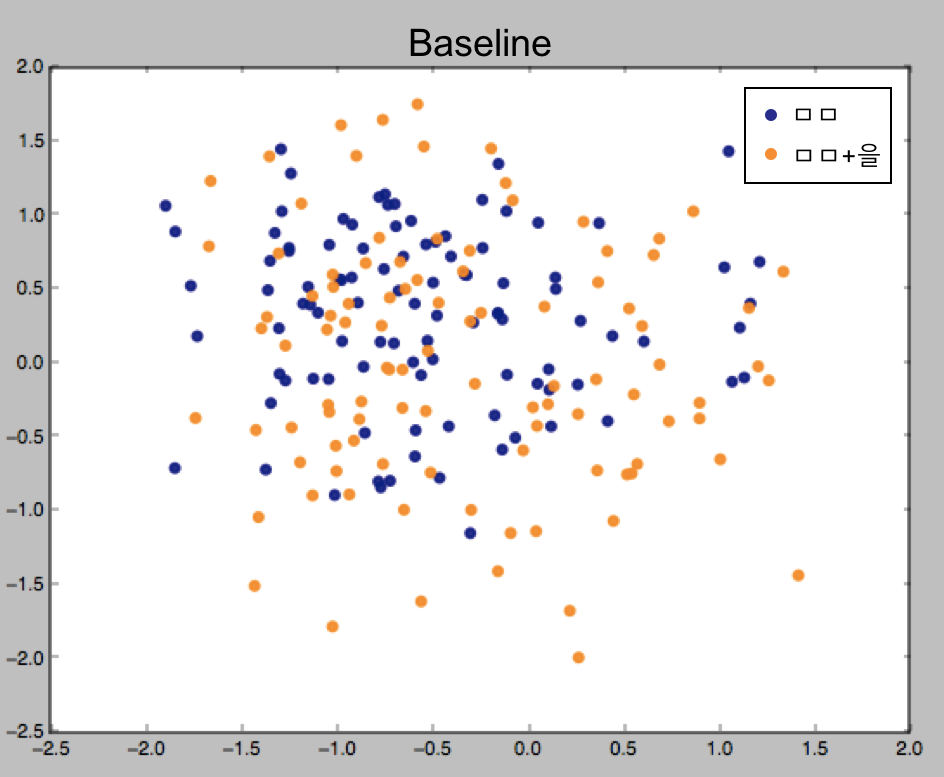}
  \end{subfigure}
  \adjustbox{minipage=2em,valign=c}{\subcaption{}\label{sfig:pca-b}}%
  \begin{subfigure}[c]{\dimexpr0.9\linewidth-1em\relax}
    \centering
    \includegraphics[width=\textwidth,valign=c]{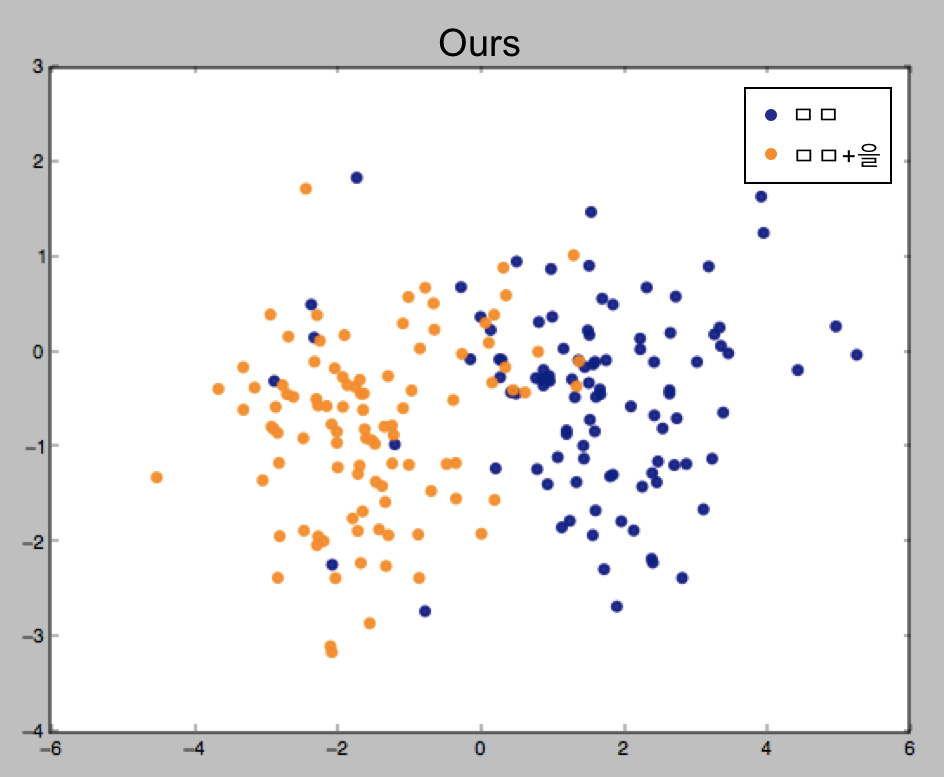}
  \end{subfigure}
  \caption{PCA projections of vector representation of 100 randomly sampled pairs of word. Each pair is composed of a word and the same word with postposition. In (b), our model shows that words forming the discriminative parallel clusters against postposition-combined-words.\label{fig:pca}}
\end{figure}

To compare how models learn the morphological characteristics, we randomly sampled hundred words and the same words combined with certain postposition(`\begin{CJK}{UTF8}{mj}을\end{CJK}', \emph{eul}) from the training data.
The graph in Figure \ref{fig:pca} shows this result more clearly.
We can observe that words forming the discriminative parallel clusters against postposition-combined-words while the baseline doesn't.

\section{Conclusion}

We present a syllable-based word representation model experimented with Korean, which is one of morphologically rich languages.
Our model keeps the characteristics of Skip-gram models, in which word representation learns from context words.
It also takes into account the morphological characteristics by sharing parameters between the words that contain common syllables.
We demonstrate that our model is competitive on quantitative evaluations.
Furthermore, we show that the model can handle OOV words, and capture morphological relationships.
As a future work, we have a plan to expand our model so that it can utilize overall information extracted from words, morphemes and characters.

\section*{Acknowledgments}
This work was supported by the National Research Foundation of Korea(NRF) funded by the Ministry of Science, ICT and Future Planning (NRF-2016M3C4A7952587, PF Class Heterogeneous High Performance Computer Development).

\bibliography{emnlp2017}
\bibliographystyle{emnlp_natbib}

\end{document}